\documentclass[10pt,twocolumn]{article}
\usepackage[utf8]{inputenc}
\usepackage{cmap}
\usepackage[T1]{fontenc}
\input{glyphtounicode}
\usepackage[margin=0.85in,top=1.0in,bottom=1.0in]{geometry}
\usepackage{amsmath,amssymb,amsthm,mathtools}
\usepackage{bm}
\usepackage[numbers,sort&compress]{natbib}
\usepackage{hyperref}
\usepackage{booktabs}
\usepackage{graphicx}
\usepackage{enumitem}
\usepackage{xcolor}

\newcommand{\R}{\mathbb{R}}

\newcommand{\lam}{\lambda}

\newcommand{\Wc}{\|W\|_{c}}
\newcommand{\Tgrok}{T_{\mathrm{grok}}}

\title{\textbf{The Weight Norm Sets the Grokking Timescale:\\
A Causal Delay Law}}
\author{Truong Xuan Khanh$^{1}$ \quad Doan Hoang Viet$^{2}$ \quad Luu Duc Trung$^{1}$ \quad Phan Thanh Duc$^{3}$\\[2pt]
\small $^{1}$H\&K Research Studio / Clevix LLC, Hanoi, Vietnam \quad
$^{2}$Bac A Bank, Hanoi \quad $^{3}$Banking Academy of Vietnam\\
\small \texttt{khanh@clevix.vn}}
\date{}

\begin{document}
\maketitle

\begin{abstract}
We show that the weight norm causally controls the \emph{timescale} of grokking, reconciling two opposing
accounts: that grokking occurs at a critical weight norm, and that the norm is not the operative variable.
On modular arithmetic, a network first memorizes by raising its weight norm and then relaxes under weight
decay; under free dynamics generalization emerges as the norm reaches a sharply concentrated value $\Wc$
($1$--$2\%$ across a two-fold range of training fractions, nearly unchanged across a ten-fold range of
learning rates), matching recent threshold reports. We then intervene: a matched-counterfactual clamp
holding $\|W\|=\rho\,\Wc$ throughout training shows the network groks \emph{at whatever norm it is held at}
--- no single norm is required --- with the time to grok growing as an exponential delay law
$T_{\mathrm{grok}}\propto e^{\alpha\rho}$ ($R^2>0.99$). Across five tested moduli (four usable for the fit)
this is a \emph{scaling law}: one shared exponent $\alpha\approx7.5$ collapses the delay ($R^2=0.996$), the
norm entering only through its value relative to $\Wc(p)$. Holding the norm above $\Wc$ thus delays grokking
rather than preventing it --- an apparent ``prevention'' under a fixed budget is the finite-budget tail of
this law --- and the norm dominates the grokking time ($\approx\!19\times$) over the learning rate
($\approx\!2\times$). This reconciles the camps: the concentrated $\Wc$ is the norm the relaxation
\emph{reaches} on its natural timescale, not a hard gate, so grokking is seen above and below it; the result
is a causal complement to norm-separation delay theory, whose closed-form delay is logarithmic in
the norm ratio under free contraction. The effect is altered by \emph{LayerNorm}, which decouples the
weight-norm scale from the function. On an un-normalized attention model the delay law recurs with its own
exponent ($\alpha_2\approx15$, $R^2=0.999$; $\approx2\times$ the MLP), so the \emph{form} of the law is
cross-architectural while its rate is not. A concentrated, regularization-set grokking norm also recurs on a
non-Fourier task (sparse parity), although the above-norm delay does not fully transfer.
\end{abstract}


\section{Introduction}
Grokking --- delayed generalization long after a network has fit its training set --- has become a
testbed for understanding when and why neural networks generalize~\citep{power2022grokking}. Two influential
accounts have emerged. The \emph{weight-norm} account, exemplified by \emph{Omnigrok}~\citep{liu2022omnigrok},
holds that generalization is governed by the parameter norm: small-norm regions generalize and weight decay
drives the network there. The \emph{circuit} account~\citep{nanda2023progress} shows mechanistically that
the network gradually builds Fourier-feature ``circuits'' that implement the arithmetic. These have largely
been studied separately and qualitatively.

This paper makes the weight-norm account \emph{sharp, quantitative, and causal}, and connects it to the
circuit account. The literature is currently divided: one line reports a concentrated weight-norm threshold
for grokking \citep{liu2022omnigrok,systematic2026grokking,neuralcollapse2026}, while another argues the
norm is correlative or not the operative variable at all
\citep{golechha2024progress,minegishi2023grokking,notsawo2025euclidean,nonstationarity2025}. Our central
aim is to reconcile this disagreement. We record the weight norm at grokking under free dynamics and find
it concentrated at $\Wc$ (in agreement with the threshold camp); we then intervene with a
matched-counterfactual clamp that holds the norm at a chosen multiple of $\Wc$ throughout training, and find
that the norm does not gate grokking at a fixed value but sets its \emph{timescale} --- a causal delay law.
This dissolves the disagreement: $\Wc$ is the norm the free relaxation reaches on its natural timescale, not
a value grokking requires. This is also consistent with a possible division of labor between the two
accounts --- circuit formation governing \emph{what} is learned and norm relaxation \emph{when} it
generalizes --- though our evidence for the link is the temporal alignment of the two, not a direct
manipulation of the circuits.

\paragraph{Contributions.}
\begin{enumerate}[leftmargin=*, itemsep=1pt, topsep=2pt]
\item \textbf{A concentrated grokking norm under free dynamics.}
  Trained freely, the model groks when the weight norm reaches a sharply concentrated value $\Wc$: for fixed
  task size and weight decay, $\Wc$ varies by only $1$--$2\%$ across a two-fold range of training fractions
  and is nearly invariant across a ten-fold range of learning rates (\S\ref{sec:critnorm}). Concurrent
  studies report the same rate-vs-threshold dissociation \citep{systematic2026grokking,neuralcollapse2026};
  we corroborate it and then show, causally, that it is a \emph{rate} phenomenon.
\item \textbf{The weight norm causally sets the grokking timescale (a delay law).}
  A matched-counterfactual clamp that holds $\|W\|=\rho\,\Wc$ throughout training reveals three things
  (\S\ref{sec:causal}): (i) the network groks at \emph{whatever} norm it is held at --- there is no single
  norm grokking requires; (ii) over $\rho\in[0.85,1.15]$ the time to grok follows a clean exponential delay
  law $T_{\mathrm{grok}}\propto e^{\alpha\rho}$ ($R^2>0.99$); (iii) holding the norm above $\Wc$ therefore
  \emph{delays} grokking rather than preventing it --- an apparent ``prevention'' under a fixed budget is the
  finite-budget tail of this exponential. The norm is the dominant control of grokking time
  ($\approx\!19\times$ across the range), with the learning rate a secondary, dissociable modulator
  ($\approx\!2\times$).
\item \textbf{The delay is a scaling law with a shared exponent across task sizes.}
  Repeating the clamp across five tested moduli $p$, the four usable sizes collapse onto a single exponential
  in the \emph{relative} norm $\rho=\|W\|/\Wc(p)$: one shared exponent $\alpha\approx7.5$ fits them with
  $R^2=0.996$ (per-$p$ CV $4.1\%$ over the tested $1.7\times$ size range), establishing a causal scaling law
  rather than a per-task fit (\S\ref{sec:scaling}). We report its limits honestly --- sub-exponential saturation beyond $\rho\approx1.25$
  and a minimum-norm floor at the smallest $p$ (which is why it is excluded from the fit) --- and relate it to norm-separation delay theory
  \citep{truong2026normsep}: that theory predicts a \emph{logarithmic} free-training
  delay via norm contraction, whereas our clamp blocks contraction and is a continuous-dose version of the
  norm-freeze causal test, confirming the norm as the operative variable (we do not claim a quantitative
  match of the exponent to a closed-form prediction).
\item \textbf{Recurrence across architectures, and the role of normalization.}
  The delay law recurs on a second, un-normalized attention model with its own exponent
  ($\alpha_2\approx15$, $R^2=0.999$; grok-at-held-norm and high-norm saturation as on the MLP), and is
  altered by LayerNorm, which
  decouples the weight-norm scale from the function; only the functionally relevant unembedding norm stays
  concentrated. The weight norm controls the timescale when it sets the function scale (\S\ref{sec:universal}).
\item \textbf{Task generality on a non-Fourier task.}
  On sparse parity, grokking again occurs at a concentrated, weight-decay-set, learning-rate-invariant norm,
  and the norm again acts through the rate rather than a hard threshold (\S\ref{sec:generality}).
\item \textbf{A temporal link to the circuit account.}
  Fourier-feature concentration rises in the same norm-relaxation window. This temporal alignment is
  consistent with --- but does not establish --- norm relaxation organizing the feature formation of the
  circuit account; we do not intervene on the circuits directly, and we make the connection only at the
  correlational level.
\end{enumerate}

\section{Related Work}

\textbf{Grokking: phenomenology and mechanism.}
Grokking was identified by \citet{power2022grokking} on small algorithmic tasks, where test accuracy rises
abruptly long after the training set is fit. \citet{nanda2023progress} reverse-engineered the learned
solution on modular arithmetic as a small set of Fourier-frequency circuits implementing a trigonometric
identity, and introduced progress measures (e.g.\ a restricted loss) that improve smoothly before the
visible jump; \citet{gromov2023grokking} gave analytic constructions of such solutions, and
\citet{barak2022hidden} showed on sparse parity that hidden progress accumulates throughout the plateau.
\citet{varma2023circuit} framed grokking as competition between a memorizing and a generalizing circuit of
differing parameter efficiency, with weight decay tipping the balance. These works establish \emph{what}
the network computes and that progress is gradual; our focus is the scalar control variable --- the weight
norm --- and its causal role in triggering the transition.

\textbf{A concentrated weight-norm threshold: an emerging, but purely observational, consensus.}
A second line attributes grokking to the parameter norm and the optimizer's implicit bias.
\citet{liu2022omnigrok} (Omnigrok) argued that generalization is organized by the weight norm: a
generalizing region exists at moderate norm, weight decay or rescaled initialization moves the network into
it, and with weight decay as the only regularizer the norm relaxes as $\|w(t)\!\approx\!e^{-\gamma t}\|w_0\|$
toward a target $w_c$ in time $\propto 1/\lambda$. Related accounts cast the jump as a lazy-to-rich
transition \citep{kumar2023grokking}, or as a late-phase implicit-bias dichotomy \citep{lyu2023dichotomy},
connected to the classical max-margin bias of gradient descent \citep{soudry2018implicit}. Two concurrent
empirical studies sharpen the role of regularization in a way closely related to ours, and we are careful to
delineate what they establish from what we add. \citet{systematic2026grokking} run a factorial sweep over
depth, architecture, activation, and weight decay on modular addition, finding that grokking is driven by
the interaction of optimization and regularization rather than by architecture --- weight decay is the
dominant control, with a narrow ``Goldilocks'' regime, and the Transformer--MLP gap nearly vanishes
($1.11\times$ delay) under matched hyperparameters; in a single weight-norm experiment (one seed per
configuration) they report that the RMS parameter norm at grokking is concentrated to $0.0219\pm0.0032$
(CV $14.5\%$) across five width-512 models, with ReLU and GELU grokking at \emph{identical} norms despite a
$6.9\times$ difference in delay. A parallel study of neural-collapse dynamics
\citep{neuralcollapse2026} reports the same rate-vs-threshold dissociation for a \emph{different} observable
--- the penultimate-layer \emph{feature} norm in image classification --- with a concentrated onset value
(within-pair CV $<8\%$) that is largely invariant to training conditions while training sets only the
\emph{rate} of approach, again with a Goldilocks zone. Both studies are \emph{observational} on the
norm--time relation: each records the norm at which the transition occurs and varies hyperparameters around
it; neither holds the norm fixed, measures a delay as a function of a controlled norm, or reports a scaling
law. Our observational findings (\S\ref{sec:critnorm}) corroborate this
rate-vs-threshold dissociation and tighten the concentration to CV $1$--$2\%$ by conditioning on task and
$\lambda$, adding invariance to learning rate and to the grokking threshold, and a modulus scaling
$\Wc\!\propto\!p^{0.38}$. We regard the concentrated threshold itself as a \emph{corroborated} phenomenon
rather than a novel one; our distinct contribution begins where these studies stop --- we test the threshold
\emph{causally} by clamping the norm and reading off the resulting delay (\S\ref{sec:causal}), and show the
response is a quantitative, cross-task delay law (\S\ref{sec:scaling}) rather than another observational
characterization. (The cross-architecture comparison above is at the natural operating point; our
$\alpha_2\!\approx\!2\alpha$ for the un-normalized transformer measures a different quantity, the
sensitivity of the delay to a \emph{held} norm, so the two are consistent.)

\textbf{Challenges to weight-norm causality.}
A vigorous counter-current argues the weight norm is \emph{not} the operative variable.
\citet{golechha2024progress} reports grokking on real-world data (MNIST, IMDb) outside the expected norm
range --- even without weight decay, with the norm increasing --- concluding weight norms are correlative,
not causal. \citet{minegishi2023grokking} show a sparse ``grokking ticket'' generalizes faster than a
\emph{norm-matched} dense network, so the norm is not sufficient. \citet{notsawo2025euclidean} demonstrate
that grokking can be driven by regularization toward properties other than the $\ell_2$ norm (sparsity,
low rank), that the $\ell_2$ norm is not a reliable proxy --- it often grows while the model generalizes ---
and that depth alone can induce grokking without explicit regularization. \citet{nonstationarity2025}
argue that what drives the transition is the \emph{effective learning rate} --- the ratio of parameter norm
to update norm --- rather than the parameter norm on its own, and show that deliberately raising it
accelerates feature learning (and mitigates primacy bias under nonstationarity); further work induces
grokking with increasing norms and no weight decay \citep{datafallsshort2025}. We take this critique seriously and do not claim the norm is universally
causal. Instead, our results delimit \emph{when} the norm is causal: it controls the grokking timescale in
settings where it sets the function scale, and that control weakens or disappears exactly where these works
operate --- under normalization (\S\ref{sec:universal}), when regularization targets a non-$\ell_2$ property,
or on a task whose solution is not pinned to a single norm (\S\ref{sec:generality}). To separate the norm
\emph{state} from the effective learning rate, our intervention holds the norm fixed while the optimizer runs (without
resetting AdamW moments) and measures a dose-response across clamp levels at fixed learning rate
(\S\ref{sec:causal}).

\textbf{Interventions on the norm, and the role of normalization.}
The interventional evidence closest to ours comes from two different settings, and in each the intervention
differs from our sustained clamp in a way that turns out to be decisive. In the neural-collapse study,
\citet{neuralcollapse2026} rescales the penultimate \emph{feature} norm to $0.3\times$ or $3.0\times$ its
natural value at the start of the collapse phase and then releases it; the feature norm self-corrects back to
the same value from either direction, with collapse time essentially unchanged ($p>0.2$), and the authors
read this as evidence that the threshold is a gradient-flow \emph{attractor rather than a cause}. This is a
one-shot rescale-and-release, and its null result is the same one we obtain for the grokking weight norm in
our own one-shot control (\S\ref{sec:causal}, ``Sustained, not instantaneous''): a single rescale washes out
because the optimizer re-equilibrates the norm, leaving the grokking time within $\sim\!15\%$ of the free
control. Far from pre-empting our finding, that null result \emph{motivates} our design --- the effect
appears only under a \emph{sustained} clamp that holds the norm away from its attractor throughout training,
which is what produces the exponential dose--response. The two studies are thus complementary across both
domain (feature norm in neural collapse vs.\ weight norm in grokking) and intervention type (transient
rescale vs.\ sustained hold). Closest to our weight-norm setting, \citet{verma2026regimes} runs a
matched-control intervention on a LayerNorm
transformer: re-initializing attention heads changes post-grokking head differentiation, while a
\emph{matched weight-norm clip} does not, isolating the effect to head structure rather than weight
magnitude. This null result for norm-clipping is consistent with --- and explained by --- our finding
that LayerNorm decouples the weight-norm scale from the function, so that clamping the total norm should have
little effect there; our gate appears precisely in the \emph{un-normalized} setting (\S\ref{sec:universal}).
The decoupling itself has a formal basis: for a final normalization layer the radial gradient at the output
vanishes, so the loss cannot be reduced by rescaling the last hidden state, whereas without it cross-entropy
pushes representation norms up \citep{gatednorm2026}; high-norm directions act on logits only through the
final LayerNorm \citep{gurnee2024confidence}, and LayerNorm's scale-invariance is classical
\citep{zhang2019rmsnorm}. Concurrently, \citet{yildirim2026geometric} relates normalization, the unembedding
matrix, and logit scale to grokking onset via a bounded-geometry intervention. Our contribution in this
setting is to show which norm remains concentrated under LayerNorm (only the functionally relevant
unembedding norm) and that the norm's control of the grokking timescale reappears once normalization is removed --- a grokking-specific
instantiation of the general decoupling these works establish.

\textbf{Relationship to our prior work.}
A companion theoretical paper from our group derives a norm-separation delay law from first principles,
predicting that the \emph{free-training} grokking delay grows logarithmically in the norm ratio as the norm
contracts to threshold \citep{truong2026normsep}. The present paper differs in object and method: it
intervenes directly on the scalar weight-norm \emph{state} (rather than deriving free dynamics) and maps the
conditions under which that state is causal for the onset of grokking. Where that theory predicts the
free-training delay as logarithmic in the norm ratio, our pinned-norm
intervention measures the complementary regime in which contraction is blocked, finding an exponential
dependence on the held relative norm. We build on, and cite, that work rather than restating it.

\textbf{Grokking and emergence as phase transitions.}
Abrupt capability jumps in training and at scale invite a statistical-physics reading. Emergent abilities
\citep{wei2022emergent} were argued by \citet{schaeffer2023emergent} to be partly an artifact of
discontinuous metrics, sharpening the need for well-defined order parameters and finite-size analysis;
singular learning theory treats learning as passing through phases of differing degeneracy
\citep{watanabe2009algebraic}. Within this view our contribution is methodological as well as empirical: we
identify a scalar control parameter --- the weight norm --- and measure its causal, quantitative effect on
the timescale of the transition, treating grokking as a phenomenon to be measured rather than merely
analogized.

\textbf{Double descent.}
Epoch-wise and model-wise double descent \citep{belkin2019reconciling,nakkiran2021deep} are distinct
generalization transitions in which test error is non-monotone near the interpolation threshold. We use
double descent as a contrast (\S\ref{sec:discussion}): in a linear model it is analytically a saddle-node
bifurcation with a different critical exponent, placing it in a different universality class from the
grokking transition studied here.

\section{Setup}
\label{sec:setup}

\textbf{Task and data.} We study modular addition $f(a,b)=(a+b)\bmod p$ for prime $p$. Each example is the
token sequence $[a,b,{=}]$ over a vocabulary of size $p{+}1$ (the $p$ residues plus an ``$=$'' token), with
target the residue $f(a,b)$. The dataset is the full set of $p^2$ pairs; a fraction $\alpha$ is sampled per
seed as the training set and the remainder held out. The training fraction $\alpha$ is our control
parameter; the modulus $p$ (dataset size $p^2$) is our system size.

\textbf{Model.} Unless noted, the network is a two-layer MLP over learned token embeddings: an embedding
matrix $E\in\R^{p\times d}$, the concatenation of the two operand embeddings ($\in\R^{2d}$), a hidden layer
$W_1\in\R^{2d\times H}$ with GELU, and an output $W_2\in\R^{H\times p}$; biases carry no weight decay. We
use $d=128$, $H=256$. A one-layer multi-head attention model (token and positional embeddings,
self-attention, an MLP block, and an unembedding, predicting at the final position) is used for the
architecture study (\S\ref{sec:universal}), in two variants --- with LayerNorm and without --- to isolate
the role of normalization.

\textbf{Optimization.} We train full-batch with AdamW ($\beta_1{=}0.9$, $\beta_2{=}0.999$,
$\epsilon{=}10^{-8}$), learning rate $r$, and decoupled weight decay $\lam$ applied to weight matrices only;
parameters are initialized at scale $1/\sqrt{\text{fan-in}}$. Reference hyperparameters are $r=10^{-3}$,
$\lam=1$ unless swept.

\textbf{Observables.} On a geometric grid of training steps and across $12$--$16$ seeds we record train and
test accuracy; the total weight norm
\begin{equation}
\|W\| = \big(\|E\|_F^2 + \|W_1\|_F^2 + \|W_2\|_F^2\big)^{1/2},
\end{equation}
the Omnigrok control variable; and the embedding's Fourier concentration, the Gini coefficient of its
token-frequency power spectrum $P_k=\sum_{j}\,|\hat{E}_{k,j}|^2$ (DC term removed, normalized to a
distribution), which is low for an unstructured embedding and high once a few Fourier modes dominate. We
define the memorization time $T_{\mathrm{mem}}$ as the first step with train accuracy $\ge 0.99$, the
grokking time $\Tgrok$ as the first step with test accuracy $\ge 0.9$ (robustness to this threshold is
examined in \S\ref{sec:critnorm}), and the critical norm $\Wc=\|W\|$ at $\Tgrok$.

\textbf{Modulus sweep.} For the observational scaling of the critical norm (\S\ref{sec:critnorm}) we sweep
$p\in\{29,41,59,79,97\}$.

\textbf{Intervention (matched counterfactual).} For the causal test (\S\ref{sec:causal}) the random seed is
derived from the base configuration only, \emph{excluding} the intervention, so the control and every
intervention share identical initialization, data split, and pre-intervention trajectory. The norm
\emph{clamp} projects the weights after each step so that $\|W\|=\rho\,\Wc$ for a chosen multiple $\rho$,
holding the norm at a multiple of the critical value; a one-shot variant applies the projection once at a
fixed step $t_{\mathrm{int}}$ (after memorization, before grokking).

\textbf{Reproducibility.} All experiment runners write atomically with per-configuration resume and store
raw trajectories and intermediate snapshots (including periodic embedding matrices) for later analysis;
code and data are released.

\section{The Critical Weight Norm}
\label{sec:critnorm}
Recording $\|W\|$ at the grokking step reveals a sharp regularity: within a fixed architecture, optimizer,
task size, and weight-decay strength, grokking occurs at a critical norm $\Wc$ concentrated to within a few
percent, varying only with weight decay and system size. We emphasize that this section is observational;
its causal status is tested in \S\ref{sec:causal}.

\begin{figure*}[t]
\centering
\includegraphics[width=0.92\textwidth]{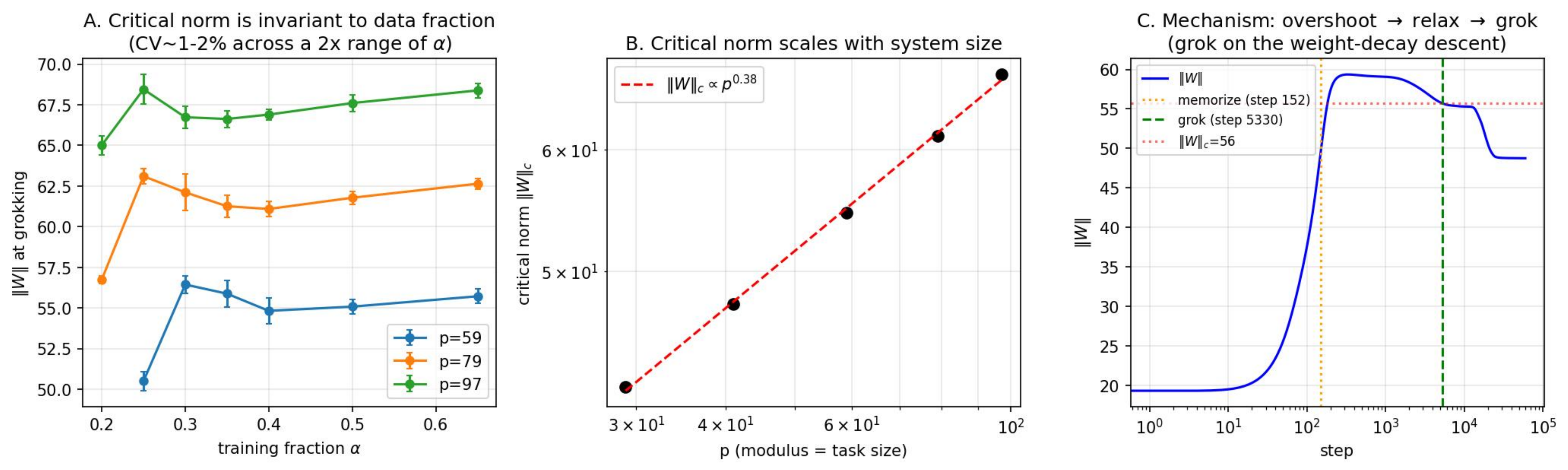}
\caption{\textbf{A critical weight norm at grokking (observational).}
\textbf{(A)} $\|W\|$ at grokking is concentrated and nearly invariant to the training fraction $\alpha$
(CV $1$--$2\%$) at each $p$. \textbf{(B)} It scales with system size, $\Wc\propto p^{0.38}$.
\textbf{(C)} The norm overshoots during memorization, then weight decay relaxes it; grokking occurs on the
descent as $\|W\|$ approaches $\Wc$, in the same window as Fourier-feature formation.}
\label{fig:critnorm}
\end{figure*}

\paragraph{Concentrated, and stable across data fraction and learning rate.}
At fixed $(p,\lam,r)$, $\|W\|$ at grokking is nearly constant across the training fraction $\alpha$: for
$p=59$ it is $\{56.4,55.9,54.8,55.1,55.7\}$ at $\alpha=\{0.30,0.35,0.40,0.50,0.65\}$ --- a coefficient of
variation of $1$--$2\%$ across a two-fold range of data (Fig.~\ref{fig:critnorm}A). At sufficient weight
decay it is also nearly independent of the learning rate: at $\lam=3$, $\Wc=\{48.9,48.7,49.1\}$ across a
ten-fold range of $r$ (CV $<1\%$), even though the grokking time changes seven-fold. The learning rate thus
changes the \emph{time to reach} $\Wc$ much more than the value of $\Wc$, separating a tunable relaxation
rate from a comparatively stable threshold; this is consistent with the $(r\lam)$-dependence of the
grokking time.

\paragraph{An internal threshold, not a training-time artifact.}
$\Wc$ is robust to the accuracy threshold defining grokking: recomputing it at test-accuracy thresholds
$0.8$, $0.9$, $0.95$ changes $\Wc$ by less than $2\%$ (e.g. $p=59\!:\,55.0/54.6/54.2$;
$p=97\!:\,68.0/67.1/66.4$), with the data-fraction stability unchanged at every threshold. Because accuracy
rises steeply at the transition while the norm relaxes slowly, the norm barely moves over the accuracy
window $0.8$--$0.95$: $\Wc$ marks an internal state, not the placement of a threshold.

\paragraph{Set by weight decay; scales with system size.}
$\Wc$ is not a universal constant. It shifts with weight decay --- $\Wc\approx 49,55,63$ at $\lam=3,1,0.3$,
consistent with a weight-decay equilibrium norm --- and scales with system size as
$\Wc(p)=\{42,48,55,61,67\}$ for $p=\{29,41,59,79,97\}$, a power law $\Wc\propto p^{0.38}$
(Fig.~\ref{fig:critnorm}B). The claim is one of threshold structure within a fixed
$(p,\lam,\text{architecture})$ family, not of a single universal value.

\paragraph{Temporally aligned with Fourier-feature formation.}
The norm overshoots during memorization (peaking near step $340$ in Fig.~\ref{fig:critnorm}C), after which
weight decay relaxes it; grokking occurs on the descent as the norm approaches $\Wc$. The embedding's
Fourier concentration rises in the same window (norm overshoot $\to$ feature formation $\to$ grokking),
suggesting that norm relaxation toward $\Wc$ \emph{may organize} the feature formation of the circuit
account. The trajectory shows temporal alignment; the next section establishes the causal role of the norm.

\section{A Causal Test: the Weight Norm Sets the Grokking Timescale}
\label{sec:causal}
The correlations above admit two readings. Either the norm \emph{controls} grokking, or it is merely a
by-product of the same training dynamics. We discriminate them with a controlled intervention. The answer is
sharper than either the threshold or the no-effect view: the norm controls \emph{when} grokking happens,
through a clean delay law.

\paragraph{Design.}
We use a matched counterfactual: the initialization, data split, and pre-intervention trajectory are
identical across conditions (the random seed depends only on the base configuration, not on the
intervention or the learning rate), so any difference is attributable to the intervention alone. The
intervention \emph{clamps} the weight norm: at every step after a fixed point $t_{\mathrm{int}}$ (chosen
after memorization, before grokking) we project the weights so that $\|W\|=\rho\,\Wc$, holding the norm at a
chosen multiple of the critical value without resetting the optimizer moments. To separate the norm
\emph{state} from the learning rate, we run a full $\rho\times\eta$ matrix, $\rho\in\{0.85,1.00,1.15,1.30\}$
and learning rate $\eta\in\{1,2,4,8\}\times10^{-3}$, $16$ seeds each, at $p=59$, $\alpha=0.40$, $\lam=1$,
with a generous $30{,}000$-step budget so that slow conditions are observed to completion rather than
censored.

\begin{figure*}[t]
\centering
\includegraphics[width=0.92\textwidth]{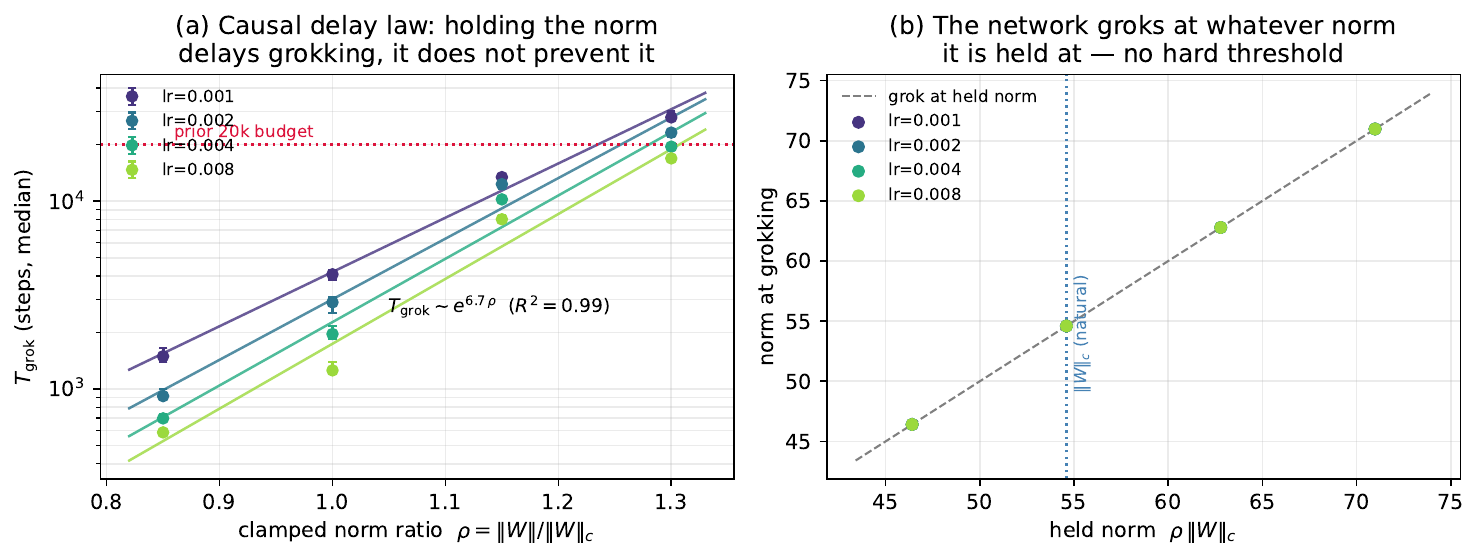}
\caption{\textbf{The weight norm causally sets the grokking timescale.}
\textbf{(A)} Holding the norm at $\rho\,\Wc$ delays grokking as a clean exponential
$\Tgrok\propto e^{\alpha\rho}$ ($R^2=0.99$); curves at four learning rates are nearly parallel (norm sets
the timescale, $\eta$ the rate). The dotted line marks a typical $20{,}000$-step budget: high-norm clamps
fall above it and look ``prevented'' under a short budget, but grok by $30{,}000$ steps.
\textbf{(B)} The network groks at \emph{whatever} norm it is held at (points lie on the diagonal), so there
is no single norm grokking requires; $\Wc$ is merely the norm the free relaxation reaches.}
\label{fig:delaylaw}
\end{figure*}

\begin{figure}[t]
\centering
\includegraphics[width=0.82\linewidth]{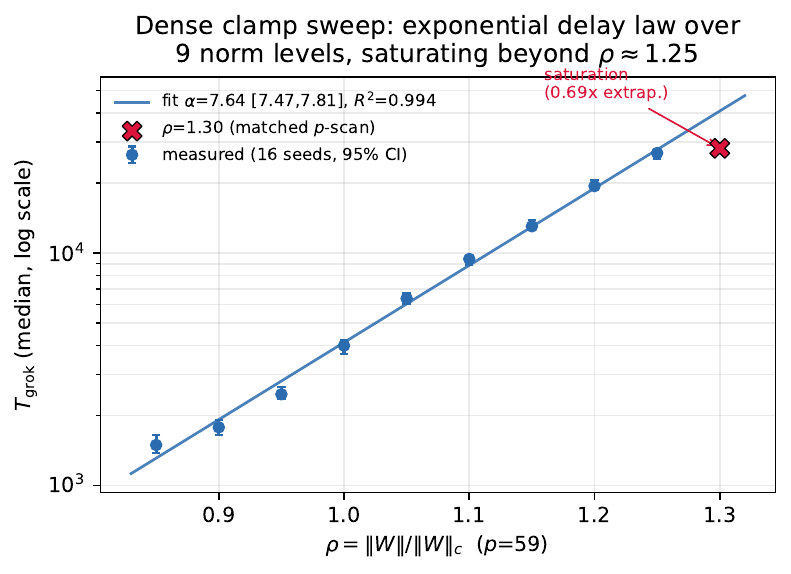}
\caption{\textbf{Dense clamp sweep ($p=59$, $16$ seeds).} The grokking delay is exponential in the held
relative norm over ten levels $\rho\in[0.85,1.25]$ ($\alpha=7.64$ [$95\%$ CI $7.47,7.81$], $R^2=0.994$);
error bars are bootstrap $95\%$ CIs on the median. The point at $\rho=1.30$ (matched $p$-scan) falls below
the extrapolation, marking the onset of sub-exponential saturation.}
\label{fig:dense}
\end{figure}

\paragraph{Result: the network groks at the held norm, after an exponentially growing delay.}
Two facts settle the causal question (Fig.~\ref{fig:delaylaw}, Table~\ref{tab:causal}). First, the network
groks at \emph{whatever} norm it is held at: across $\rho\in\{0.85,1.00,1.15,1.30\}$ the norm at grokking
equals the clamp value ($46.4,54.6,62.8,71.0$), not a fixed $\Wc$. There is no single norm that grokking
requires. Second, over the range $\rho\in[0.85,1.15]$ the time to grok grows as a clean exponential in the
held norm,
\begin{equation}
\Tgrok(\rho)\ \propto\ e^{\alpha\rho},\qquad \alpha\approx 7.4,\quad R^2>0.99,
\label{eq:delaylaw}
\end{equation}
fit independently at each learning rate (Fig.~\ref{fig:delaylaw}A); a dense $10$-level sweep below extends
this exponential through $\rho=1.25$ (Fig.~\ref{fig:dense}), beyond which (at $\rho\approx1.30$) the growth
becomes sub-exponential (\S\ref{sec:scaling}). Holding the norm above $\Wc$ thus \emph{delays} grokking rather than
preventing it: at $\rho=1.30$ ($\|W\|=71$, $30\%$ above critical) the median grokking time is
$\approx\!2.8\times10^{4}$ steps, so under a shorter budget the same condition appears to ``never''
generalize. Indeed no $\rho=1.30$ seed groks before $20{,}000$ steps, which is exactly why a fixed
$20{,}000$-step budget (and our own earlier report) read this as prevention; by $30{,}000$ steps
$12$--$16/16$ seeds grok. The ``prevention'' is the finite-budget tail of the delay law, not a separate
phenomenon.

\paragraph{Norm dominates; learning rate modulates; they dissociate.}
The norm is by far the larger lever: across the tested range it changes $\Tgrok$ by $\approx\!19\times$
(at $\eta=10^{-3}$, $\Tgrok$ rises from $1492$ at $\rho=0.85$ to $27{,}925$ at $\rho=1.30$). The learning
rate changes it only $\approx\!2\times$ (e.g.\ at $\rho=1.30$, $27{,}925\!\to\!16{,}893$ as $\eta$ goes
$1\!\to\!8\times10^{-3}$). The two axes are separable: raising $\eta$ shifts the delay-law intercept
slightly but leaves the steep $\rho$-dependence intact (the curves in Fig.~\ref{fig:delaylaw}A are nearly
parallel). The norm sets the \emph{timescale} of grokking; the learning rate sets the rate within it. This
directly addresses the view that grokking is governed by the effective learning rate rather than the norm
\citep{nonstationarity2025}: the dominant, order-of-magnitude control here is the norm state, with the
learning rate a secondary modulator.

\begin{table}[t]
\centering\small
\caption{The weight norm causally sets the grokking \emph{timescale} (matched control; $p=59$, $16$ seeds,
$30{,}000$-step budget). Median $\Tgrok$ (steps) over the $\rho\times\eta$ clamp matrix; the network groks
at the held norm in every cell. Bootstrap $95\%$ CIs on the median are tight (e.g.\ $\rho{=}1.00,\eta{=}1e{-}3$:
$4077\,[3795,4257]$; $\rho{=}1.15,\eta{=}1e{-}3$: $13426\,[13045,13817]$); per-cell CIs are in the
supplement. All cells grok $16/16$ within budget except the $\rho{=}1.30$ row at $\eta\in\{1,4\}\times10^{-3}$
($12$--$14/16$; the remaining seeds are right-censored at $30{,}000$, so those two medians are mild
underestimates), which is why \S\ref{sec:scaling} re-runs $\rho{=}1.30$ at a $160{,}000$-step budget where
all seeds grok uncensored. The apparent ``prevention'' at high $\rho$ in earlier short-budget runs is the
finite-budget tail of the exponential delay (\ref{eq:delaylaw}).}
\label{tab:causal}
\begin{tabular}{lcccc}
\toprule
held $\|W\|=\rho\Wc$ & $\eta{=}1e{-}3$ & $2e{-}3$ & $4e{-}3$ & $8e{-}3$ \\
\midrule
$0.85\,\Wc=46.4$ & 1492 & 916 & 697 & 587 \\
$1.00\,\Wc=54.6$ & 4077 & 2891 & 1961 & 1256 \\
$1.15\,\Wc=62.8$ & 13426 & 12317 & 10221 & 8007 \\
$1.30\,\Wc=71.0$ & 27925 & 23168 & 19501 & 16893 \\
\bottomrule
\end{tabular}
\end{table}

\paragraph{A dense sweep confirms the exponential and locates its edge.}
To check that the exponential is not an artifact of fitting only three norm levels, we ran a fine sweep at
$p=59$ over ten clamp levels $\rho\in\{0.85,0.90,\dots,1.30\}$ ($16$ seeds each), measuring $\Wc=54.5$ from
a free control. The network groks at the held norm in every cell, all $16/16$ seeds, uncensored
(Fig.~\ref{fig:dense}). Over the nine levels $\rho\in[0.85,1.25]$ the delay is a clean exponential,
$\alpha=7.64$ with a bootstrap $95\%$ CI of $[7.47,7.81]$ and $R^2=0.994$ --- a tight fit over a $30\times$
range of grokking times, not an underdetermined three-point line. The exponential regime is wider than our
initial estimate: it holds through $\rho=1.25$, and only at $\rho=1.30$ does the delay fall clearly below the
extrapolation ($\approx\!0.69\times$, matching the $p$-scan value of \S\ref{sec:scaling}), marking the onset
of the sub-exponential saturation. There is a mild downward curvature at the very bottom ($\rho\le0.90$),
where the held norm approaches the minimum functional norm (\S\ref{sec:scaling}).

\paragraph{Is it the norm, or just stronger regularization?}
A natural objection is that constraining the norm low is simply stronger regularization, which already
accelerates grokking. We address this with a matched control. Training freely at a range of weight decays
$\lam\in\{1,2,3,5,8\}$ traces a regularization frontier on the $(\|W\|\text{ at grokking},\Tgrok)$ plane
(lower equilibrium norm, faster grokking). The below-clamp points lie \emph{on} this frontier: the
below-direction speed-up is not distinguishable from stronger regularization, and we do not claim it as a
separate effect. The \emph{above}-norm delay, however, is not reproducible by regularization: since weight
decay can only \emph{lower} the equilibrium norm, no free-$\lam$ run sits at the slow, high-norm states
($\|W\|=63$--$71$) that the clamp holds, so the long delays at $\rho>1$ are a property of the norm state
that no weight-decay setting reaches. An \emph{Omnigrok-style} low-norm initialization (rescaling the
initial weights, then training freely) is washed out: the norm relaxes back toward $\Wc$ and grokking occurs
on the free timescale rather than being accelerated by the low start --- consistent with the delay law,
since only a \emph{sustained} hold (not the initial value) changes the held norm the dynamics run at.

\paragraph{Sustained, not instantaneous.}
A \emph{one-shot} rescale at a single step produces only a weak effect, because the optimizer
re-equilibrates the norm and the kick washes out; the large effect requires the \emph{continuous} clamp.
Quantitatively, a one-shot global rescale to $\rho\,\Wc$ that is then released leaves the grokking time
nearly flat across $\rho\in[0.70,1.30]$ (median $T_{\mathrm{grok}}$ moves only $1.2\times$, staying within
$\sim\!15\%$ of the free control), whereas a \emph{sustained} hold of the same norm spans $7.3\times$ over
the narrower range $\rho\in[0.90,1.15]$. A one-shot $0.8\times$ scaling of any single parameter group
(embedding, hidden, or output) likewise leaves $T_{\mathrm{grok}}$ within $\sim\!3\%$ of the control. What
matters is the norm state the network is held in throughout the relaxation, not a one-time crossing or a
transient perturbation of the optimizer --- exactly as the delay law (\ref{eq:delaylaw}) requires.

\paragraph{The total norm, not inter-layer structure.}
The clamp is a strong, sustained constraint (a per-step projection in the spirit of Omnigrok's rescaling),
not a subtle perturbation. It rescales all weights by a single global factor, so it fixes the \emph{total}
magnitude while leaving the optimizer free to redistribute norm across layers. It does so: under the held
clamp the per-layer norm fractions (embedding, hidden, output) still drift over training by as much
(maximum shift $\approx0.06$) as in free training ($\approx0.06$), even though the total is pinned. The
delay is therefore a response to the total norm \emph{value}, not to a freezing of the relative inter-layer
structure. (We did not run a \emph{sustained} per-layer pin, so we make no claim about how independently
holding each layer's norm would compare.) Overall the matrix establishes that the norm \emph{state}
causally sets the grokking timescale, dominant over the learning rate, and that holding the norm high
delays rather than prevents grokking. The intervention does not by itself establish that norm relaxation
causes the Fourier features specifically; that link remains temporal (\S\ref{sec:critnorm}).

\begin{figure}[t]
\centering
\includegraphics[width=\linewidth]{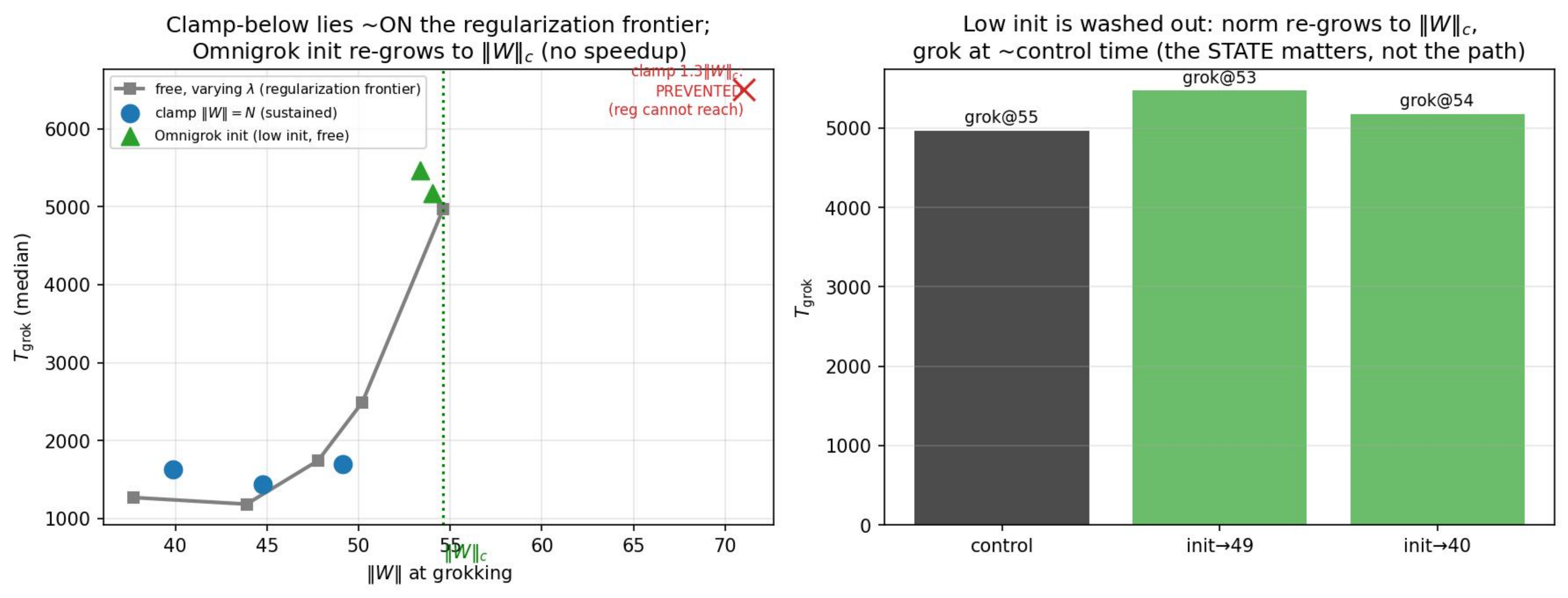}
\caption{\textbf{The norm state, not just regularization.}
\textbf{(Left)} Free training at varying weight decay traces a $(\|W\|\text{ at grok},\Tgrok)$ frontier;
the below-direction norm clamps lie on it (their speed-up is regularization), but the slow high-norm states
the clamp holds ($\rho>1$) sit off the frontier, at norms no weight decay reaches, so the above-norm delay
is a property of the norm state and not of regularization.
\textbf{(Right)} An Omnigrok-style low-norm initialization is washed out: the norm relaxes back toward
$\Wc$ and grokking occurs on the free timescale, so only a \emph{sustained} hold changes the timescale.}
\label{fig:normctl}
\end{figure}

\section{A Scaling Law for the Grokking Delay}
\label{sec:scaling}
The single-task delay law (\ref{eq:delaylaw}) raises the question that separates a genuine law from a fitted
curve: is the exponent $\alpha$ a property of one task size, or does the same law govern grokking across
task sizes? We answer this by repeating the clamp intervention across five moduli $p\in\{31,41,53,59,71\}$.
For each $p$ we first measure $\Wc(p)$ from a free control and then clamp the norm at $\rho\,\Wc(p)$ for
$\rho\in\{0.85,1.00,1.15\}$ (and $\rho=1.30$ at $p\in\{41,59\}$), $12$ seeds per cell. The smallest modulus
$p=31$ does not grok when clamped at $\rho=0.85$ --- the held norm falls below the minimum functional norm
for the task (a floor we discuss in \S\ref{sec:scaling}) --- so we fit the exponent on the four larger
moduli and treat $p=31$ as the low-end boundary of the usable range.

\begin{figure}[t]
\centering
\includegraphics[width=\linewidth]{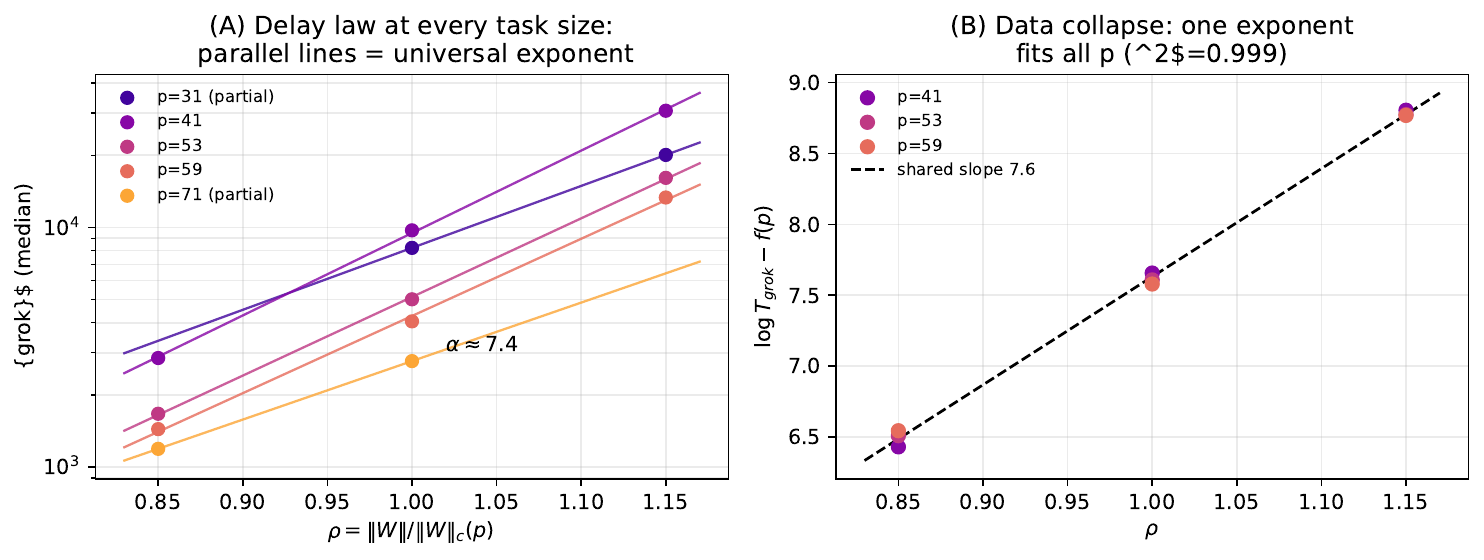}
\caption{\textbf{A scaling law for the grokking delay.}
\textbf{(A)} Across task sizes $p$, holding the norm at $\rho\,\Wc(p)$ delays grokking exponentially over
$\rho\in[0.85,1.15]$ (lines; near-parallel $\Rightarrow$ shared exponent $\alpha\approx7.5$). The
$\times$ markers at $\rho=1.30$ fall below the extrapolation: the law saturates beyond $\rho\approx1.15$.
\textbf{(B)} Data collapse: with one shared exponent and a per-$p$ offset $f(p)$, all task sizes fall on a
single line ($\alpha=7.49$, $R^2=0.996$).}
\label{fig:scaling}
\end{figure}

\paragraph{A shared exponent across the tested task sizes.}
Two results establish a scaling law (Fig.~\ref{fig:scaling}). First, at every task size the network again
groks at \emph{exactly} the held norm (norm at grokking $=\rho\,\Wc(p)$ in every cell; each cell groks
$12/12$ seeds with the slowest seed well within budget, so none of these medians is censored), and the delay
follows the same exponential form over $\rho\in[0.85,1.15]$ with a shared exponent: fitting each
$p$ independently gives $\alpha=7.92,7.56,7.42,7.07$ for $p=41,53,59,71$ ($R^2\ge0.986$), a mean of
$\bar\alpha=7.49$ (coefficient of variation $4.1\%$). The point estimates decrease monotonically with $p$,
but this drift is within seed noise: the bootstrap $95\%$ CIs of the four per-$p$ exponents all overlap (in
$[7.06,7.38]$), and replacing the single shared slope by four free per-$p$ slopes (three extra parameters)
reduces the residual sum of squares only from $0.054$ to $0.037$, which no model-selection criterion
justifies. The data are therefore consistent with a single exponent. Second, the curves \emph{collapse}: a
global model $\log\Tgrok=\alpha\,\rho+f(p)$ with one shared exponent fits all four task sizes with a
bootstrap estimate $\alpha=7.49$ ($95\%$ CI $[7.16,7.69]$) and $R^2=0.996$ (Fig.~\ref{fig:scaling}B). A
single exponent across the tested $1.7\times$ range of $p$ is the signature of a scaling law rather than a
per-task coincidence: the weight norm sets the grokking timescale through the \emph{same} exponential
dependence on the norm \emph{relative} to its task-specific critical value $\rho=\|W\|/\Wc(p)$. (This
relative normalisation is also the right one empirically: re-collapsing against the absolute norm $\|W\|$
degrades the fit to $R^2=0.988$, whereas $\rho$ and other $p$-rescalings match within $0.001$.) We still
describe the exponent as shared over the \emph{tested} finite-size range rather than proven universal, since
four task sizes cannot exclude a slow drift outside this range. The critical value itself scales smoothly,
$\Wc(p)\propto p^{0.44}$, consistent with the observational scaling of \S\ref{sec:critnorm}, and the
functionally relevant embedding norm scales similarly ($\|E\|_c\propto p^{0.47}$).

\paragraph{Honest limits of the exponential: high-norm saturation and a low-norm floor.}
The exponential is not unbounded. At $\rho=1.30$ the measured delay falls well \emph{below} the
extrapolation of the $[0.85,1.15]$ fit (at $p=41$, $3.9\times10^{4}$ steps versus a predicted
$1.0\times10^{5}$, a ratio of $0.39$; at $p=59$, ratio $0.71$), so the growth is sub-exponential beyond
$\rho\approx1.15$ --- the exponential law has an upper edge, and we report it as holding over
$\rho\in[0.85,1.15]$ rather than indefinitely. These high-$\rho$ cells are run at a $160{,}000$-step budget
and are \emph{not} censored: all $12/12$ seeds grok, with the slowest seed ($\le\!4.7\times10^{4}$ steps)
finishing more than threefold below the budget, so the measured delays are full distributions rather than
budget-limited lower bounds. At the opposite end, the smallest task ($p=31$) clamped \emph{below} its
critical norm
($\rho=0.85$, held $\|W\|=34$) does not grok within budget --- it memorizes but its test accuracy creeps up
only slowly --- suggesting a minimum functional norm below which the solution cannot form efficiently at
small $p$; at larger $p$ the same $\rho=0.85$ lies above this floor and accelerates grokking as expected.
Both edges sharpen rather than weaken the central finding: across the regime where a single weight norm
cleanly sets the function scale, the grokking delay obeys one exponential scaling law in the relative norm.

\paragraph{Relation to delay theory.}
First-passage / norm-separation accounts of grokking
\citep{truong2026normsep} model free training as a crossing problem: weight decay
contracts the parameter norm $V_t=\|\theta_t\|^2$ exponentially until it reaches an architecture-dependent
threshold $V_\star$, at which the validation transition occurs, giving a delay that is
\emph{logarithmic} in the norm ratio, $T_{\mathrm{grok}}-T_{\mathrm{mem}}\approx(2\kappa\,\eta\lambda)^{-1}\log(V_{\mathrm{mem}}/V_\star)$.
Our clamp does not measure that logarithmic contraction law --- it is a different experiment. By pinning the
norm we \emph{block} the contraction route and ask how long grokking takes when the norm cannot reach
$V_\star$ from above; this is a continuous-dose version of the norm-freeze causal intervention those accounts
use to establish necessity (freezing the norm prevents grokking). Our dose-response confirms the central
claim of that picture --- that the norm is the operative crossing variable --- and adds a quantitative law
for the pinned regime: the delay grows \emph{exponentially} in the pinned relative norm, steeply enough that
a fully frozen high norm reproduces the ``never groks within budget'' of the freeze intervention. The
exponential pinned-norm law is thus complementary to, not the same as, the logarithmic free-contraction law:
it characterises the regime where the norm-contraction route is unavailable and generalization must instead
be reached at fixed norm. We do not claim a quantitative match between the measured exponent
$\alpha\approx7.5$ and a closed-form theoretical prediction; deriving the pinned-norm exponent from the
joint norm--angle dynamics is left to future work.

\section{Architecture Dependence and the Role of Normalization}
\label{sec:universal}
Does the critical-norm phenomenon recur beyond the MLP? We test it on a second architecture --- a one-layer
attention model --- in two variants, and find that the answer turns on a single architectural choice,
normalization. We use this to map where the norm controls the timescale rather than to claim broad
universality.

\begin{figure}[t]
\centering
\includegraphics[width=\linewidth]{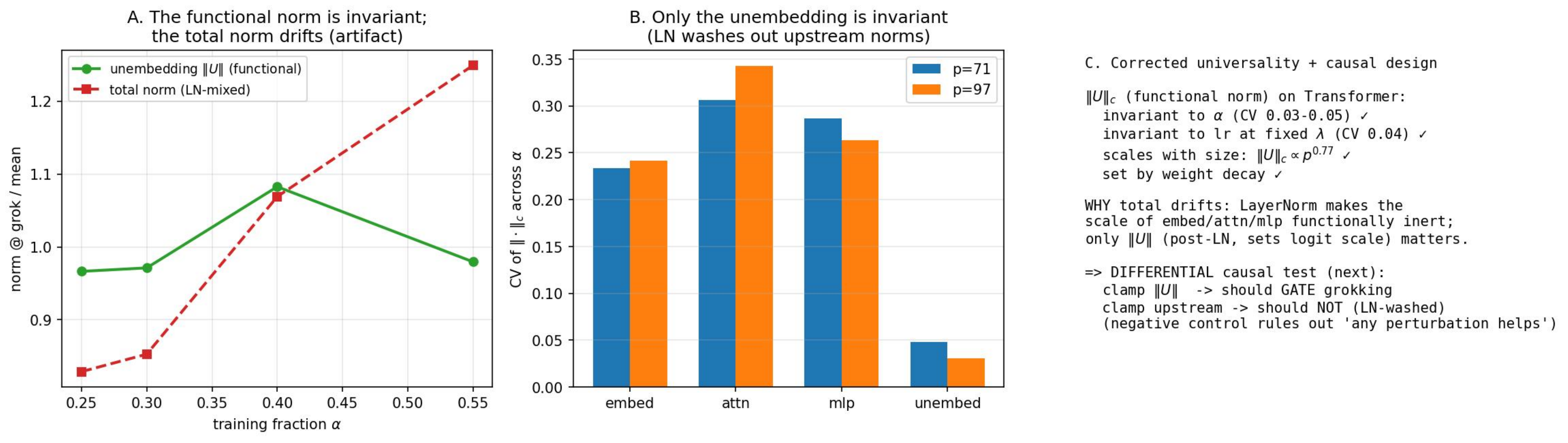}
\caption{\textbf{On a LayerNorm Transformer, the threshold replicates on the functional norm.}
\textbf{(A)} The unembedding norm $\|U\|$ at grokking is invariant to the training fraction while the total
norm drifts. \textbf{(B)} Across-$\alpha$ coefficient of variation by parameter group: only the
unembedding is concentrated (LayerNorm washes out the upstream norms).}
\label{fig:funcnorm}
\end{figure}

\paragraph{The functionally relevant norm replicates the threshold.}
On a one-layer attention model \emph{with} LayerNorm, the \emph{total} weight norm at grokking is not
concentrated: it drifts with the training fraction and learning rate (CV $0.15$--$0.17$ across $\alpha$).
This is an artifact of the wrong observable. LayerNorm renormalizes the activations, so the scale of the
embedding, attention, and MLP weights (which sit before a normalization) is functionally inert and free to
drift. The one norm that sets the function is the \emph{unembedding} $U$ (after the final LayerNorm, it
sets the logit scale). Measured on $\|U\|$, the threshold returns: $\|U\|$ at grokking is concentrated
(CV $0.03$--$0.05$ across $\alpha$, $0.04$ across a tenfold $r$ range at fixed $\lam$), is set by weight
decay, and scales with system size as $\|U\|_c\propto p^{0.77}$ (Fig.~\ref{fig:funcnorm}). The
critical-norm threshold thus replicates on the Transformer, but on the functionally relevant norm.

\begin{figure}[t]
\centering
\includegraphics[width=0.86\linewidth]{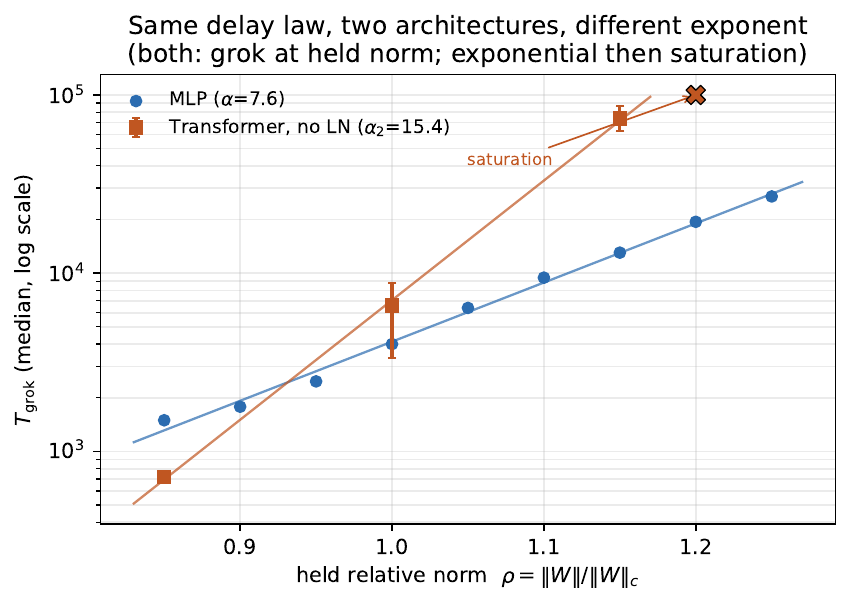}
\caption{\textbf{The delay law transfers to a second architecture with its own exponent.} On the
un-normalized one-layer attention model (orange, $16$ seeds, bootstrap $95\%$ CIs), the grokking delay is
exponential in the held relative norm, $\Tgrok\propto e^{\alpha_2\rho}$ with $\alpha_2=15.45$
($R^2=0.999$ over $[0.85,1.15]$), saturating at $\rho=1.20$ --- the same shape as the MLP (blue,
$\alpha\approx7.6$) but $\approx2\times$ steeper. Both architectures grok at the held norm.}
\label{fig:tfdelay}
\end{figure}

\paragraph{The norm's control of the timescale is broken by LayerNorm and restored without it.}
We then clamp by group. Clamping any single group on the LayerNorm model --- including $\|U\|$ --- moves
grokking only weakly and non-specifically (holding $\|U\|$ above its grokking value delays grokking up to
$\sim\!2\times$, and clamping an upstream group has a comparable effect), because the residual stream mixes
contributions and LayerNorm rescales them, so no single weight norm cleanly sets the function. Removing
LayerNorm changes this entirely. On the \emph{un-normalized} one-layer attention model, the total weight
norm again sets the function scale, and its control of the grokking timescale reappears as a full delay law
(Fig.~\ref{fig:tfdelay}). With the critical norm measured at $\Wc\approx37$ from a free control, an
extended-budget clamp sweep groks at \emph{exactly} the held norm in every cell (norm at grokking
$=\rho\,\Wc$: $31.4,36.9,42.4,44.3$ for $\rho=0.85,1.00,1.15,1.20$), all $16/16$ seeds and uncensored, with
median $\Tgrok=718,\,6608,\,73966,\,99920$. Crucially, the $\rho=1.15$ case --- which under a short
$30{,}000$-step budget plateaued below $0.9$ and looked like \emph{prevention} --- groks at
$\approx7.4\times10^{4}$ steps once the budget is extended, confirming a delay rather than a hard block. The
delay is exponential in the held norm, $\Tgrok\propto e^{\alpha_2\rho}$ with $\alpha_2=15.45$ ($95\%$ CI
$[14.84,15.98]$, $R^2=0.999$ over $\rho\in[0.85,1.15]$), and saturates beyond: at $\rho=1.20$ the measured
delay is $0.64\times$ the exponential extrapolation, the same exponential-then-saturation shape seen on the
MLP (\S\ref{sec:causal}--\ref{sec:scaling}). The exponent is about twice the MLP's ($\alpha\approx7.5$): the
\emph{form} of the delay law --- grok at the held norm, exponential rise, high-norm saturation --- transfers
across architectures, while its \emph{rate} is architecture-specific. The dose-response is sharp and
bidirectional, mirroring the MLP.

\paragraph{Interpretation.}
The weight norm controls the grokking timescale whenever it sets the scale of the function --- on the MLP
and on the un-normalized attention model --- and LayerNorm is precisely the operation that breaks this link
by renormalizing activations and decoupling the weight-norm scale from the function. This shows the causal
mechanism \emph{recurs} on a second, un-normalized architecture, and delimits where it does not apply
(normalized networks), where the concentrated norm survives only on the post-normalization readout. We
therefore speak of structural recurrence under an explicit condition --- that a single weight norm sets the
function scale --- rather than of architecture-independent universality. A methodological corollary: in a
normalized network, ``the weight norm'' must mean the functionally relevant post-normalization norm, not
the raw total.

\begin{figure}[t]
\centering
\includegraphics[width=\linewidth]{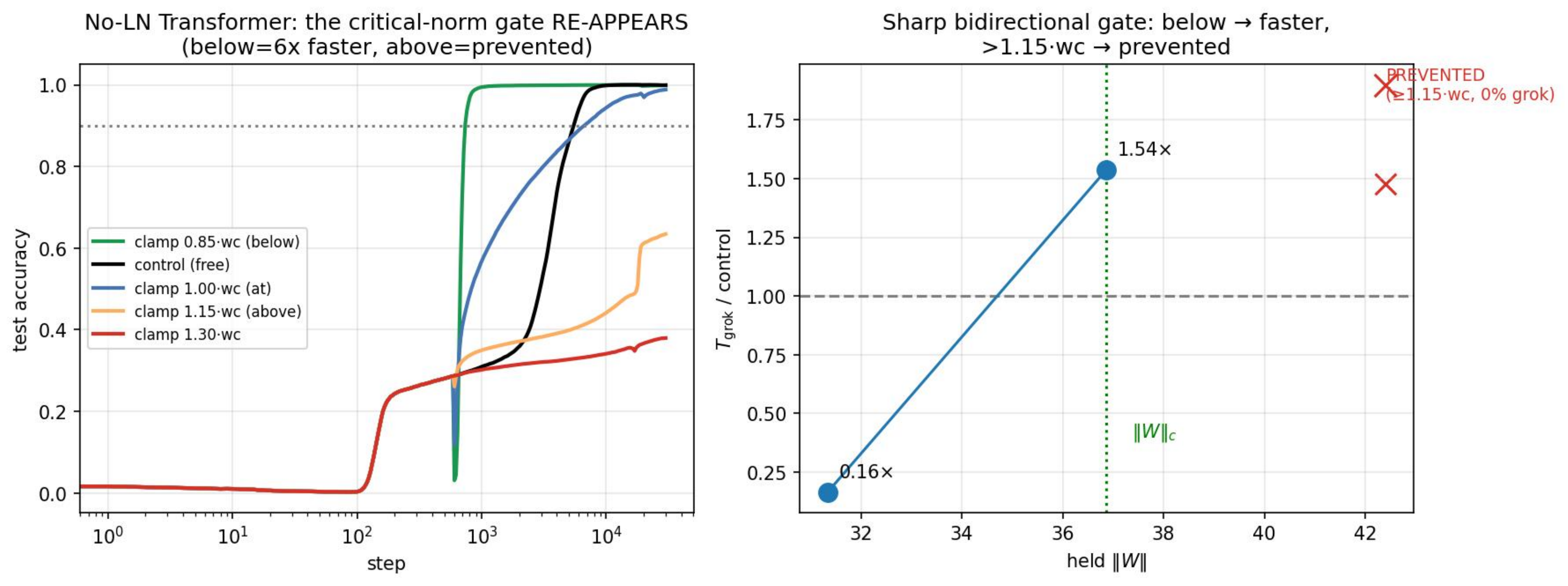}
\caption{\textbf{Without LayerNorm, the norm's control of the timescale reappears.}
\textbf{(A)} Test-accuracy trajectories on the un-normalized attention model: held below $\Wc$ groks
$6\times$ faster, control intermediate, held above $\Wc$ delayed (trajectories shown at the original
$30{,}000$-step budget). \textbf{(B)} Sharp bidirectional dose-response. At extended budget every held-norm
case groks at the held norm and follows the exponential delay law of Fig.~\ref{fig:tfdelay}; the above-norm
plateaus here are its finite-budget tail, not prevention.}
\label{fig:noln}
\end{figure}

\section{Generality: a Non-Fourier Task}
\label{sec:generality}
Modular arithmetic has cyclic, Fourier structure, which raises the question of whether the critical norm is
an artifact of that structure. We test on $k$-sparse parity --- the canonical non-Fourier grokking
task~\citep{barak2022hidden}, where the label is the product of $k$ relevant bits among $n$ and the
solution is combinatorial feature selection rather than a Fourier circuit --- using an MLP (no
normalization, no embedding), so the total weight norm is again the functionally relevant norm.

Two parts of the phenomenon transfer. First, grokking occurs. Second, the grokking norm is again a
concentrated, weight-decay-set, learning-rate-invariant target: at a fixed task and weight decay, varying
the learning rate over a fourfold range changes the grokking norm by only $\approx3\%$ while the grokking
time varies $\sim\!4\times$, and weight decay sets the grokking norm along a clean monotone frontier (as on
modular arithmetic). What does \emph{not} transfer is the regularization-unreachable above-norm delay. The
grokking norm here drifts with the amount of training data (it is not data-invariant), and the norm clamp
lies \emph{on} the weight-decay frontier in both directions: unlike the modular case, the high-norm regime
is reachable by reducing weight decay, so clamping the norm high reproduces what weak regularization already
does, rather than holding the network in a slow, high-norm state that no weight decay can reach. A plausible
reason is structural: sparse parity admits many generalizing solutions spread over a range of norms, whereas
modular addition is solved by a near-unique Fourier construction pinned to a particular scale, so only the
latter has a high-norm regime that regularization cannot otherwise enter (we offer this as a hypothesis, not
a tested claim).

We therefore distinguish two claims by their scope. The \emph{concentrated, regularization-set grokking
norm} is task-general (modular arithmetic and sparse parity; MLP and, on the functional norm, the
Transformer). The \emph{regularization-unreachable above-norm delay} --- the slow, high-norm states the
clamp holds that no weight decay reaches, together with the initialization washout --- is specific to
settings where a single weight norm sets the function scale (modular arithmetic and the un-normalized
attention model). Table~\ref{tab:generality} summarizes.

\begin{table}[t]
\centering\small
\caption{Which properties transfer across task and architecture.}
\label{tab:generality}
\begin{tabular}{@{}p{0.62\linewidth}cc@{}}
\toprule
property & mod.\ add. & parity \\
\midrule
grokking occurs                          & yes & yes \\
norm at grok set by weight decay         & yes & yes \\
norm at grok invariant to learning rate  & yes & yes \\
norm at grok invariant to data amount    & yes & \textbf{no} \\
reg.-unreachable above-norm delay        & yes & \textbf{no} \\
\bottomrule
\end{tabular}
\end{table}

\begin{figure}[t]
\centering
\includegraphics[width=\linewidth]{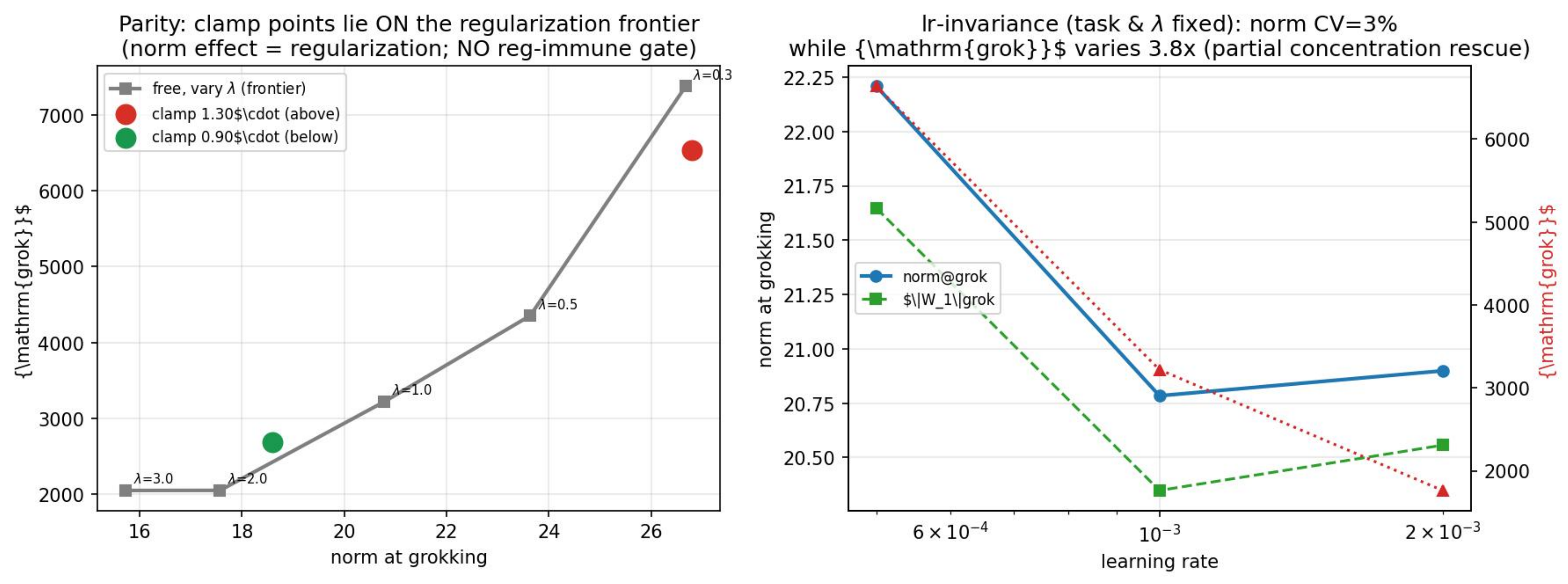}
\caption{\textbf{Sparse parity (non-Fourier).}
\textbf{(Left)} The norm clamps lie on the weight-decay frontier in both directions (no
regularization-unreachable high-norm state, unlike modular arithmetic). \textbf{(Right)} At fixed task and weight decay,
the grokking norm is nearly invariant to learning rate ($\approx3\%$) while the grokking time varies
$\sim\!4\times$ --- a concentrated, regularization-set target.}
\label{fig:parity}
\end{figure}

\begin{table*}[t]
\centering\small
\caption{Claims and evidence strength. We separate what is corroborated, what is causal, and what is
scoped, to make the reach of each claim explicit.}
\label{tab:claims}
\begin{tabular}{@{}p{0.34\textwidth}p{0.40\textwidth}l@{}}
\toprule
claim & evidence & status \\
\midrule
critical norm under free dynamics & observational sweeps (\S\ref{sec:critnorm}) & corroborated \\
norm controls grokking time & matched clamp matrix (\S\ref{sec:causal}) & causal \\
exponential delay over $\rho\!\in\![0.85,1.25]$ & dense 10-level $\rho$ sweep, $R^2{=}0.994$ & established (MLP) \\
shared exponent across $p$ & multi-$p$ collapse, $R^2{=}0.996$ & finite-range scaling \\
saturation beyond $\rho\!\approx\!1.25$ & $\rho{=}1.30$ cells, uncensored & observed \\
norm vs.\ projection & $\rho{=}1.00$ control & causal \\
second-architecture delay law & un-normalized attention, $\alpha_2{\approx}15$, $R^2{=}0.999$ & causal (confirmed) \\
sparse-parity transfer & non-Fourier MLP & partial \\
relation to delay theory & complements log contraction law & qualitative \\
\bottomrule
\end{tabular}
\end{table*}

\section{Discussion}
\label{sec:discussion}
Throughout, we keep claims separated by evidence strength --- what is observationally corroborated, what is
established causally, and what holds only over the tested finite range --- as summarized in
Table~\ref{tab:claims}.

\textbf{Reconciling the threshold and ``norm-is-not-causal'' accounts.} Our central result dissolves a
standing disagreement. One line reports a concentrated weight-norm threshold for grokking
\citep{liu2022omnigrok,systematic2026grokking,neuralcollapse2026}; another argues the norm is correlative
or not the operative variable \citep{golechha2024progress,minegishi2023grokking,notsawo2025euclidean,nonstationarity2025}.
The clamp matrix shows both are seeing facets of a \emph{rate} law: the network groks at whatever norm it is
held at, and the held norm sets the grokking time exponentially (\ref{eq:delaylaw}). The concentrated $\Wc$
seen under free training is therefore the norm the weight-decay relaxation \emph{reaches} on its natural
timescale --- which is why it looks like a threshold --- and not a value grokking requires, which is why
grokking is observed above and below it. This complements norm-separation delay theory
\citep{truong2026normsep}, which predicts the \emph{free-training} delay as
logarithmic in the norm ratio from an exponential norm-contraction argument: that theory explains why free
training groks on a finite timescale (the norm contracts to threshold), while our pinned-norm intervention
--- a continuous-dose form of its norm-freeze causal test --- isolates the norm as the operative variable
and measures the (exponential) cost of being held away from threshold.

\textbf{A bridge between two accounts.} The weight-norm and circuit accounts appear as levels of one
transition: weight decay drives the norm down, and feature formation --- and generalization --- follow on
the timescale the norm sets. The intervention shows the norm causally sets the timescale; the temporal
alignment suggests, but does not prove, that it organizes the features.

\textbf{Normalization decouples the norm from the function.} The architecture comparison
(\S\ref{sec:universal}) gives a mechanistic account of \emph{when} the picture applies. In an un-normalized
network the weight norm sets the function scale, and its control of the timescale is sharp; LayerNorm
renormalizes the activations, so the scale of upstream weights no longer affects the function and no single
weight norm cleanly controls the transition. The concentrated norm survives on a LayerNorm model only on
the post-normalization readout (the unembedding). We thus read the weight-norm account as conditional on the
norm being functionally meaningful --- a condition normalization layers are designed to relax.

\textbf{Contrast with double descent.} Epoch-wise double descent is a related but mechanistically distinct
transition: in a multi-task linear model it is a saddle-node bifurcation with transition-time exponent
$\delta=1/2$, whereas the grokking time here grows much more steeply with the control parameter. We draw the
contrast at the level of mechanism (a saddle-node escape versus a relaxation-set timescale) rather than
asserting distinct universality classes from exponents, given the limited size range.

\textbf{Generality and its limits.} Across a non-Fourier task (sparse parity) and a second architecture
(attention), the concentrated, weight-decay-set, learning-rate-invariant grokking norm recurs, while the
regularization-unreachable above-norm delay (slow high-norm states no weight decay reaches; initialization
washout) is confined to settings where a single weight norm sets the function scale. We therefore make the
broad claim only for the concentrated grokking norm and the delay law, and scope the
regularization-unreachable part explicitly.

\textbf{Is there an upper norm above which grokking never occurs?} Within the surveyed range
$\rho\in[0.85,1.30]$ grokking always occurs --- it is only delayed --- and the delay grows
\emph{sub}-exponentially at the top of this range rather than diverging, so we see no evidence of a finite
norm at which the delay becomes infinite. We cannot, however, rule out an upper threshold $\rho^{*}$ at
norms well above those tested: our intervention establishes a delay law and its saturation, not the absence
of a hard ceiling at extreme norm. We leave the existence of such a $\rho^{*}$ open.

\textbf{Is it the norm, or the projection?} Because the clamp re-projects the weights every step, one might
worry the repeated projection itself --- rather than the norm value --- causes the delay. The $\rho=1.00$
arm is the control for this: it applies the identical per-step projection but at the network's natural norm,
and it groks on essentially the free-training timescale (no delay). Delay appears only as the projected norm
departs from $\Wc$, and scales with that departure, so the operative variable is the held norm value, not
the act of projecting.

\textbf{Limitations.} The critical norm depends on weight decay and system size, so its concentration is one
of structure within a fixed family, not of an absolute value. The clamp is a strong, sustained constraint
rather than a subtle perturbation (it changes magnitude while preserving direction). The exponential delay
law is established on the MLP over $\rho\in[0.85,1.25]$ and, at an extended budget, on the un-normalized
attention model over $\rho\in[0.85,1.15]$ ($\alpha_2=15.45$, $R^2=0.999$), each saturating beyond; we did
not extend the sweep to a third architecture. The architecture study uses
one-layer attention models. On the LayerNorm model we establish that no single weight norm controls the
timescale but do not exhaustively rule out a multi-norm account. The scaling exponents are measured over a
modest size range and reported as suggestive.

\section{Conclusion}
The weight norm causally sets the timescale of grokking. Under free training the norm at grokking is sharply
concentrated at $\Wc$ (in agreement with recent threshold reports), set by weight decay and scaling with
system size. But a matched-counterfactual clamp shows this is not a hard threshold: the network groks at
whatever norm it is held at, and the time to grok grows as a clean exponential in the held norm
($T_{\mathrm{grok}}\propto e^{\alpha\rho}$, $R^2=0.99$). Holding the norm above $\Wc$ therefore delays
grokking rather than preventing it --- an apparent ``prevention'' under a fixed budget is the finite-budget
tail of this law --- and the norm dominates the grokking time ($\approx\!19\times$) over the learning rate
($\approx\!2\times$), with the two dissociable. This reconciles the threshold and ``norm-is-not-causal''
accounts: the concentrated $\Wc$ is the norm the relaxation reaches on its natural timescale, not a value
grokking requires, so grokking is seen above and below it. The picture recurs on an un-normalized attention
model and is altered by LayerNorm, which decouples the weight-norm scale from the function; on sparse parity
the concentrated, weight-decay-set, learning-rate-invariant grokking norm recurs. The result is a causal
complement to norm-separation delay theory --- whose closed-form delay is logarithmic in the norm ratio
under free contraction --- and recasts grokking as a transition whose
timescale is set by the relaxation of a functionally meaningful weight norm.
Code and data: \url{https://github.com/ClevixLab/critical-norm-grokking}. We release the runners and
analyzers, and the per-seed metrics for the full $\rho\times\eta$ clamp matrix, the multi-$p$ scan, and the
dense $\rho$ sweep, so that every fit, table, and figure here can be reproduced end to end.

\section*{Data availability}
The code and the per-seed metrics required to reproduce every number, table, and figure in this paper are
openly available at \url{https://github.com/ClevixLab/critical-norm-grokking}. The complete raw dataset,
including the full per-step structural snapshots, is archived on Zenodo (DOI to be inserted at proof stage).

\section*{Declaration of competing interest}
The authors declare that they have no known competing financial interests or personal relationships that
could have appeared to influence the work reported in this paper.

\section*{Funding}
This research did not receive any specific grant from funding agencies in the public, commercial, or
not-for-profit sectors.

\appendix
\section{Reproducibility Checklist}
\label{app:repro}
All results are produced by the released code (\url{https://github.com/ClevixLab/critical-norm-grokking}).
Each runner writes a timestamped output directory containing its own script, per-seed metrics
(\texttt{T\_grok} for every seed, per-layer norms, norm-at-grokking), and a checkpoint enabling
resume-after-interruption; analyzers read these directories and emit the numbers quoted here. The mapping
from artifact to result is:

\begin{itemize}[noitemsep]
\item \textbf{Observational critical norm} (Fig.~\ref{fig:critnorm}, \S\ref{sec:critnorm}): free-training
sweeps over training fraction and learning rate; the per-$p$ controls also supply $\Wc(p)$.
\item \textbf{Causal clamp matrix} (Fig.~\ref{fig:delaylaw}, Table~\ref{tab:causal}, \S\ref{sec:causal}):
\texttt{run\_efflr\_dissoc} --- the $\rho\times\eta$ matrix at $p=59$, $16$ seeds, with the matched control.
\item \textbf{Dense $\rho$ sweep} (Fig.~\ref{fig:dense}, \S\ref{sec:causal}): \texttt{run\_denserho} /
\texttt{analyze\_denserho} --- ten $\rho$ levels at $p=59$, giving $\alpha=7.64$, $R^2=0.994$.
\item \textbf{Scaling collapse} (Fig.~\ref{fig:scaling}, \S\ref{sec:scaling}): \texttt{run\_delaylaw\_pscan}
/ \texttt{analyze\_delaylaw\_pscan} --- the multi-$p$ scan. The single-exponent verdict
(Table~\ref{tab:claims}) is reproduced by \texttt{analyze\_scaling\_variable} (per-$p$ $\alpha$ bootstrap
CIs, shared-vs-per-$p$ model comparison, normalisation-variable test).
\item \textbf{Second architecture} (Fig.~\ref{fig:tfdelay}, \S\ref{sec:universal}):
\texttt{run\_tfnoln\_long} / \texttt{analyze\_tfnoln\_long} --- the extended-budget un-normalized
transformer, giving $\alpha_2=15.45$, $R^2=0.999$.
\item \textbf{LayerNorm functional norm and sparse parity} (Figs.~\ref{fig:funcnorm}, \ref{fig:noln},
\ref{fig:parity}; Table~\ref{tab:generality}): the LayerNorm and parity runners in the same repository.
\end{itemize}

\small
\bibliographystyle{plainnat}

\begin{thebibliography}{99}
\bibitem[Power et al.(2022)]{power2022grokking}
A.~Power, Y.~Burda, H.~Edwards, I.~Babuschkin, and V.~Misra.
\newblock Grokking: Generalization beyond overfitting on small algorithmic datasets.
\newblock \emph{arXiv:2201.02177}, 2022.
\bibitem[Nanda et al.(2023)]{nanda2023progress}
N.~Nanda, L.~Chan, T.~Lieberum, J.~Smith, and J.~Steinhardt.
\newblock Progress measures for grokking via mechanistic interpretability.
\newblock \emph{ICLR}, 2023.
\bibitem[Gromov(2023)]{gromov2023grokking}
A.~Gromov.
\newblock Grokking modular arithmetic.
\newblock \emph{arXiv:2301.02679}, 2023.
\bibitem[Barak et al.(2022)]{barak2022hidden}
B.~Barak, B.~L. Edelman, S.~Goel, S.~Kakade, E.~Malach, and C.~Zhang.
\newblock Hidden progress in deep learning: SGD learns parities near the computational limit.
\newblock \emph{NeurIPS}, 2022.
\bibitem[Varma et al.(2023)]{varma2023circuit}
V.~Varma, R.~Shah, Z.~Kenton, J.~Kram\'ar, and R.~Kumar.
\newblock Explaining grokking through circuit efficiency.
\newblock \emph{arXiv:2309.02390}, 2023.
\bibitem[Liu et al.(2022)]{liu2022omnigrok}
Z.~Liu, E.~J. Michaud, and M.~Tegmark.
\newblock Omnigrok: Grokking beyond algorithmic data.
\newblock \emph{arXiv:2210.01117}, 2022.
\bibitem[Kumar et al.(2024)]{kumar2023grokking}
T.~Kumar, B.~Bordelon, S.~J. Gershman, and C.~Pehlevan.
\newblock Grokking as the transition from lazy to rich training dynamics.
\newblock \emph{ICLR}, 2024.
\bibitem[Lyu et al.(2024)]{lyu2023dichotomy}
K.~Lyu, J.~Jin, Z.~Li, S.~S. Du, J.~D. Lee, and W.~Hu.
\newblock Dichotomy of early and late phase implicit biases can provably induce grokking.
\newblock \emph{ICLR}, 2024.
\bibitem[Soudry et al.(2018)]{soudry2018implicit}
D.~Soudry, E.~Hoffer, M.~S. Nacson, S.~Gunasekar, and N.~Srebro.
\newblock The implicit bias of gradient descent on separable data.
\newblock \emph{JMLR}, 19(70):1--57, 2018.
\bibitem[Thilak et al.(2022)]{thilak2022slingshot}
V.~Thilak, E.~Littwin, S.~Zhai, O.~Saremi, R.~Paiss, and J.~Susskind.
\newblock The slingshot mechanism.
\newblock \emph{arXiv:2206.04817}, 2022.
\bibitem[Wei et al.(2022)]{wei2022emergent}
J.~Wei et al.
\newblock Emergent abilities of large language models.
\newblock \emph{TMLR}, 2022.
\bibitem[Schaeffer et al.(2023)]{schaeffer2023emergent}
R.~Schaeffer, B.~Miranda, and S.~Koyejo.
\newblock Are emergent abilities of large language models a mirage?
\newblock \emph{NeurIPS}, 2023.
\bibitem[Watanabe(2009)]{watanabe2009algebraic}
S.~Watanabe.
\newblock \emph{Algebraic Geometry and Statistical Learning Theory}.
\newblock Cambridge University Press, 2009.
\bibitem[Belkin et al.(2019)]{belkin2019reconciling}
M.~Belkin, D.~Hsu, S.~Ma, and S.~Mandal.
\newblock Reconciling modern machine-learning practice and the classical bias--variance trade-off.
\newblock \emph{PNAS}, 116(32):15849--15854, 2019.
\bibitem[Nakkiran et al.(2021)]{nakkiran2021deep}
P.~Nakkiran, G.~Kaplun, Y.~Bansal, T.~Yang, B.~Barak, and I.~Sutskever.
\newblock Deep double descent: Where bigger models and more data hurt.
\newblock \emph{JSTAT}, 2021(12):124003, 2021.
\bibitem[Golechha(2024)]{golechha2024progress}
S.~Golechha.
\newblock Progress measures for grokking on real-world tasks.
\newblock \emph{arXiv:2405.12755}, 2024.
\bibitem[Minegishi et al.(2023)]{minegishi2023grokking}
G.~Minegishi, Y.~Iwasawa, and Y.~Matsuo.
\newblock Grokking tickets: Lottery tickets accelerate grokking.
\newblock \emph{arXiv:2310.19470}, 2023.
\bibitem[Manir \& Rupa(2026)]{systematic2026grokking}
S.~B. Manir and A.~P. Rupa.
\newblock A systematic empirical study of grokking: Depth, architecture, activation, and regularization.
\newblock \emph{arXiv:2603.25009}, 2026.
\bibitem[Rupa(2026)]{neuralcollapse2026}
A.~P. Rupa.
\newblock Neural collapse dynamics: Depth, activation, regularisation, and feature norm thresholds.
\newblock \emph{arXiv:2604.00230}, 2026.
\bibitem[Notsawo et al.(2025)]{notsawo2025euclidean}
P.~J.~T. Notsawo, G.~Dumas, and G.~Rabusseau.
\newblock Grokking beyond the Euclidean norm of model parameters.
\newblock \emph{arXiv:2506.05718}, 2025.
\bibitem[Lyle et al.(2025)]{nonstationarity2025}
C.~Lyle, G.~Sokar, R.~Pascanu, and A.~Gy\"orgy.
\newblock What can grokking teach us about learning under nonstationarity?
\newblock \emph{Conference on Lifelong Learning Agents (CoLLAs)}, 2025. arXiv:2507.20057.
\bibitem[Singh et al.(2025)]{datafallsshort2025}
V.~Singh, E.~Belilovsky, and R.~Aljundi.
\newblock When data falls short: Grokking below the critical threshold.
\newblock \emph{arXiv:2511.04760}, 2025.
\bibitem[Verma(2026)]{verma2026regimes}
L.~Verma.
\newblock Weight decay regimes in grokking transformers: Cheap online diagnostics.
\newblock \emph{arXiv:2605.20441}, 2026.
\bibitem[Kanavalau et al.(2026)]{gatednorm2026}
A.~Kanavalau, C.~Amo~Alonso, and S.~Lall.
\newblock Gated normalization removal and scale anchoring in pre-norm transformers.
\newblock \emph{arXiv:2602.10408}, 2026.
\bibitem[Stolfo et al.(2024)]{gurnee2024confidence}
A.~Stolfo, B.~Wu, W.~Gurnee, Y.~Belinkov, X.~Song, M.~Sachan, and N.~Nanda.
\newblock Confidence regulation neurons in language models.
\newblock \emph{NeurIPS}, 2024. arXiv:2406.16254.
\bibitem[Zhang \& Sennrich(2019)]{zhang2019rmsnorm}
B.~Zhang and R.~Sennrich.
\newblock Root mean square layer normalization.
\newblock \emph{NeurIPS}, 2019.
\bibitem[Y{\i}ld{\i}r{\i}m(2026)]{yildirim2026geometric}
A.~Y{\i}ld{\i}r{\i}m.
\newblock The geometric inductive bias of grokking: Bypassing phase transitions via architectural topology.
\newblock \emph{arXiv:2603.05228}, 2026.
\bibitem[Truong et al.(2026)]{truong2026normsep}
T.~X. Khanh, T.~Q. Hoa, L.~D. Trung, and P.~T. Duc.
\newblock The norm-separation delay law of grokking: A first-principles theory of delayed generalization.
\newblock \emph{arXiv:2603.13331}, 2026.
\end{thebibliography}

\end{document}